%% file: iclr2026_conference.tex
\documentclass{article}
\usepackage{iclr2026_conference,times}

\input{math_commands.tex}

\usepackage{hyperref}
\usepackage{url}
\usepackage{latexsym}
\usepackage[T1]{fontenc}
\usepackage[utf8]{inputenc}
\usepackage{microtype}
\usepackage{inconsolata}
\usepackage{graphicx}
\usepackage{booktabs}
\usepackage{amsfonts}
\usepackage{nicefrac}
\usepackage{xcolor}
\usepackage{colortbl}
\usepackage{arydshln}
\usepackage{enumitem}
\usepackage{amsmath,amssymb,mathtools}
\usepackage{algorithm}
\usepackage{algpseudocode}
\usepackage{multirow}
\usepackage{makecell}
\usepackage{wrapfig}
\usepackage{caption}
\usepackage[most]{tcolorbox}
\tcbuselibrary{listings, breakable}
\usepackage{tikz}
\usetikzlibrary{arrows.meta, positioning, calc, fit, backgrounds}

\definecolor{retroBlue}{RGB}{68, 108, 169}
\newtcblisting{PromptBox}[2][]{
    enhanced jigsaw,
    breakable,
    listing only,
    listing options={
        basicstyle=\footnotesize\rmfamily,
        breaklines=true,
        breakatwhitespace=false,
        columns=fullflexible,
        keepspaces=true,
        aboveskip=0pt, belowskip=0pt,
        breakindent=0pt,
        breakautoindent=false,
    },
    colframe=black!55,
    colback=black!6,
    coltitle=white,
    colbacktitle=black!65,
    title={#2},
    fonttitle=\large\bfseries,
    arc=4mm,
    boxsep=3pt,
    top=2pt, bottom=2pt,
    #1
}

\graphicspath{{figures/}}

\title{ARCO: Adaptive Rubrics with Co-Evolution for Multi-Step LLM-Based Agents}

\author{Zihang Tian, Jingsen Zhang, Rui Li, Xiaohe Bo, Yuanzi Li, Xu Chen\thanks{Corresponding author.}\\
Renmin University of China\\
{\fontfamily{pcr}\selectfont zihangtian@ruc.edu.cn, xu.chen@ruc.edu.cn}
}

\iclrpreprintcopy % Uncomment for public preprint/arXiv version.
\makeatletter
\ClearShipoutPicture
\makeatother
\begin{document}

\maketitle

\input{sections/abstract.tex}
\input{sections/introduction.tex}
\input{sections/preliminary.tex}
\input{sections/method.tex}
\input{sections/experiments.tex}
\input{sections/related_work.tex}
\input{sections/conclusion.tex}
\input{sections/ethics_repro.tex}

\bibliography{bibtex,custom}
\bibliographystyle{iclr2026_conference}

\clearpage
\appendix
\input{sections/appendix_warmup.tex}
\input{sections/appendix_runtime.tex}
\input{sections/appendix_step_signal.tex}

\input{sections/appendix_case_figures.tex}
\input{sections/appendix_prompts.tex}
\end{document}

%% file: math_commands.tex
\usepackage{amsmath,amsfonts,bm}

\def\eqref#1{equation~\ref{#1}}
\def\1{\bm{1}}

\DeclareMathAlphabet{\mathsfit}{\encodingdefault}{\sfdefault}{m}{sl}
\SetMathAlphabet{\mathsfit}{bold}{\encodingdefault}{\sfdefault}{bx}{n}

%% file: sections/abstract.tex
\begin{abstract}
Reinforcement learning for multi-step LLM agents often relies on scalar rewards that indicate success but cannot explain why a trajectory is good or bad. Rubric-based rewards improve interpretability through
natural-language criteria, but existing methods share two limitations: they score \emph{at the trajectory level}, offering no guidance for individual steps; and their scorer is closed-source and static, so it cannot
adapt as the agent evolves during training. We propose \textbf{ARCO} (\textbf{A}daptive \textbf{R}ubric \textbf{CO}-evolution), which generates a \emph{per-step} rubric and predicts a rubric-conditioned
\emph{step-level} reward for each action, and continually updates this rubric model on on-policy rollouts so that its criteria and scores co-evolve with the agent's improving behavior. Across HotpotQA,
2WikiMultiHopQA, and MuSiQue with two open-source backbones, ARCO achieves the highest EM in all settings over outcome-, rubric-, and process-reward baselines, and analyses show its rubrics are step-specific, robust
to design choices, and useful for diagnosing agent behavior. Code and data are available at \url{https://github.com/zihangtian/ARCO}.
\end{abstract}

%% file: sections/introduction.tex
\section{Introduction}
\label{sec:introduction}
Reinforcement learning (RL) has become a standard approach for improving multi-step LLM-based agents on tasks such as multi-hop QA and tool use~\cite{zeng2025reinforcing,kalyan2025reinforcement,jin2025searchr1}. Most methods train with black-box scalar rewards---either trajectory-level outcome rewards or step-level process rewards---that indicate quality but cannot explain \emph{why} a trajectory is good or bad~\cite{ouyang2022training,lightman2023let,xi2025agentprm}. This opacity invites reward hacking and limits the granularity of credit assignment across steps~\cite{gao2023scaling}.

Rubric-based rewards offer a more interpretable alternative by grounding evaluation in explicit natural-language criteria~\cite{gunjal2025rubrics,gupta2025carmo,shao2025dr,liu2025openrubrics}. Yet existing rubric methods share two limitations. First, they all operate at the trajectory level: a single rubric scores the entire rollout, leaving the credit-assignment problem at individual steps untouched. Second, they rely on a frozen closed-source LLM (e.g., GPT-4) as the external judge. Whether the rubric is static~\cite{gunjal2025rubrics}, dynamically generated per trajectory~\cite{gupta2025carmo}, or evolved via an on-policy buffer~\cite{shao2025dr}, only the input \emph{text} changes while the judge's parameters are never updated.

Consider a multi-hop question: \emph{In what year was the university where Sergei Tokarev was a professor founded?} A black-box reward assigns the same score to a trajectory that luckily guesses ``1755'' and one that methodically searches for Tokarev's university before looking up its founding year. A static rubric might check ``did the agent search?''---but once the agent learns to search, its errors shift to subtler modes: targeting the wrong entity, redundant retrieval, or premature answers. A frozen judge cannot adapt its scoring to these evolving failure modes, regardless of how the rubric text is updated.

\begin{figure*}[t]
\centering
\includegraphics[
    width=\textwidth,
    trim=6pt 0pt 5pt 0pt,
    clip
]{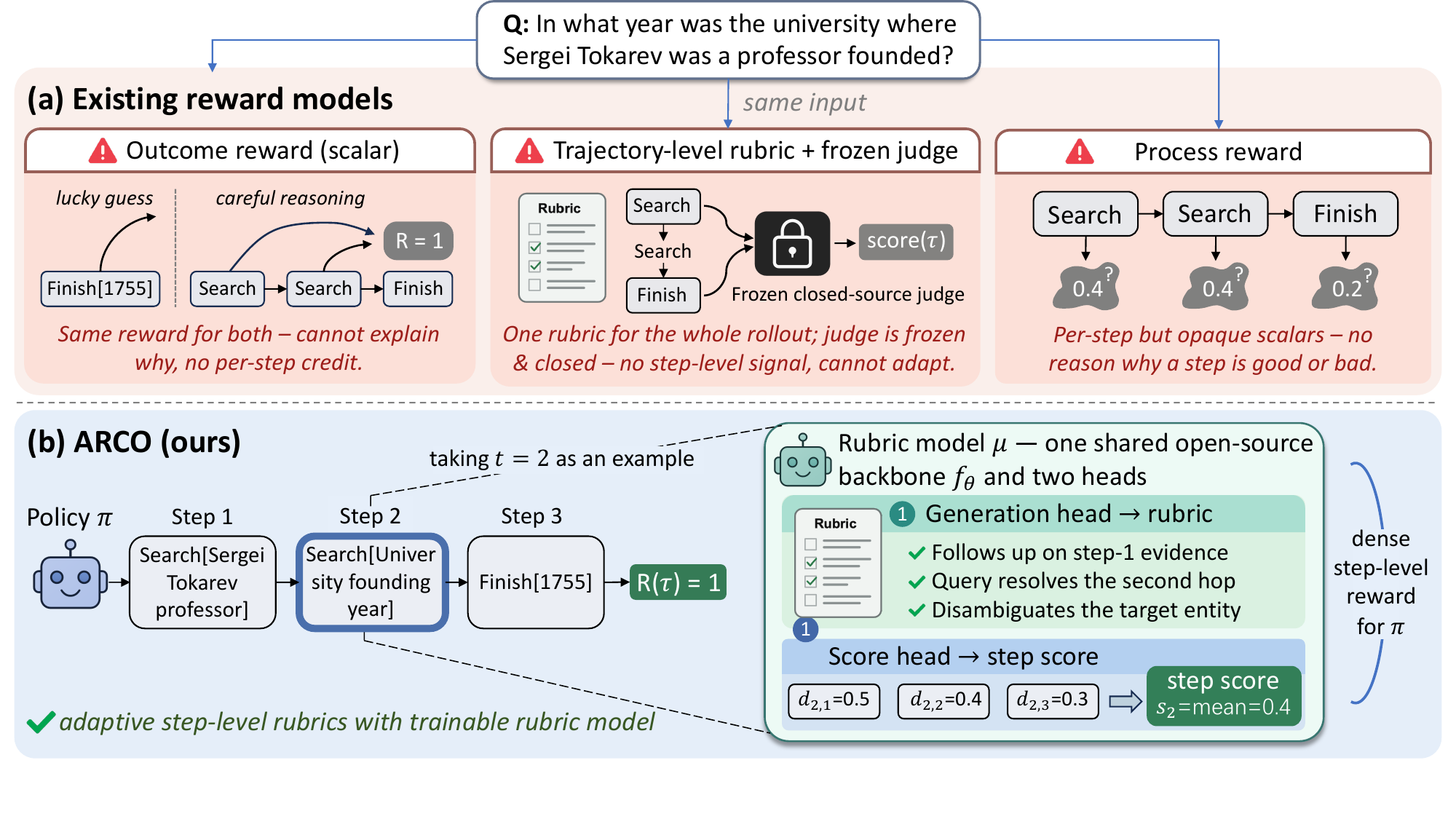}
\vspace{-30pt}
\caption{Reward signals for a multi-hop QA agent. \textbf{(a)}~Existing reward models each fall short: \emph{outcome reward} is a single scalar with no per-step credit; a \emph{trajectory-level rubric} scored by a frozen judge cannot adapt to the policy; and a \emph{PRM} gives per-step but opaque scores. \textbf{(b)}~ARCO uses a single open-source rubric model $\mu$ to generate a natural-language rubric per action and score the action against it, yielding an interpretable, step-level, adaptive reward with no external judge.}
\label{fig:architecture}
\vspace{-10pt}
\end{figure*}

We argue that an effective reward model for multi-step agents should be (i)~\emph{rubric-based}, providing interpretable criteria rather than opaque scalars; (ii)~\emph{step-level}, assigning per-action scores rather than a single trajectory score; and (iii)~\emph{co-evolutionary}, updating both the evaluation criteria and the scoring function alongside the policy at the parameter level. We propose \textbf{ARCO} (\textbf{A}daptive \textbf{R}ubric \textbf{CO}-evolution), a single trainable rubric model $\mu$ that couples two complementary functions on a shared backbone: a \emph{generation head} producing per-step natural-language criteria and a \emph{score head} predicting rubric-conditioned step rewards. Both $\mu$ and the policy $\pi$ are updated jointly on on-policy data, so the evaluation criteria and scoring function co-evolve with the agent's improving behavior. Figure~\ref{fig:architecture} contrasts ARCO with existing outcome, trajectory-rubric, and process rewards.

Our main contributions are:
\begin{itemize}[leftmargin=*,nosep]
    \item We introduce \textbf{ARCO}, an adaptive, \emph{step-level} rubric-based process reward framework in which both the natural-language rubric and the scoring function co-evolve with the policy at the parameter level, removing the dependence on a frozen closed-source judge.
     \item We propose a unified rubric model $\mu$ that couples rubric generation and action scoring into a single, jointly optimizable objective on a shared backbone so that
     the rubric and the reward are learned together rather than as two decoupled stages.
    \item We conduct extensive experiments demonstrating ARCO's effectiveness, along with comprehensive analyses that inform future work on interpretable process rewards.
\end{itemize}

%% file: sections/preliminary.tex
\section{Preliminary}
\label{sec:preliminary}

A multi-step LLM-based agent solves a task $q \in \mathcal{Q}$ by sequentially selecting actions $a_t \sim \pi(\cdot \mid h_t)$ given a trajectory prefix $h_t = (q, a_1, o_1, \ldots, a_{t-1}, o_{t-1})$, producing a trajectory $\tau$ that receives a binary outcome reward $R(\tau) \in \{0,1\}$ only at
termination. The dominant paradigm optimizes $\pi$ directly against this terminal signal with an \emph{outcome reward model} (ORM), as in search-augmented agents trained by outcome-only RL~\cite{jin2025searchr1,song2025r1searcher}; but because the reward is sparse and trajectory-level, it says nothing about
\emph{which} intermediate action helped or hurt, making credit assignment hard. To densify the signal, a \emph{process reward model} (PRM) assigns a scalar score $s_t$ to each step~\cite{xi2025agentprm}, so $\pi$ receives feedback at every action rather than only at the end; yet a PRM compresses all facets
of step quality into one opaque number, revealing neither which criterion an action violated nor why. Rubric-based process supervision makes this feedback explicit by pairing the score with natural-language criteria: at each step a generator $\rho$ produces a rubric $r_t$ (a set of evaluation criteria) and a
scorer $g$ predicts a step reward $s_t$ conditioned on it,
\begin{equation}
\label{eq:rubric_form}
r_t = \rho(h_t, a_t),\qquad s_t = g(h_t, a_t, r_t).
\end{equation}
Existing rubric methods, however, differ in three limiting design axes. \emph{Rubric adaptation}: criteria may be static (fixed before training)~\cite{gunjal2025rubrics,liu2025openrubrics}, dynamically generated per trajectory but with a frozen generator~\cite{gupta2025carmo}, or evolving at the content
level via on-policy buffer management~\cite{shao2025dr}---yet the scoring function's parameters remain frozen, and criteria are typically shared across all steps. \emph{Scoring granularity}: $g$ usually produces a single trajectory-level score rather than per-step scores
$s_t$~\cite{gupta2025carmo,gunjal2025rubrics}. \emph{Judge model}: these methods rely on closed-source LLMs (e.g., GPT) as external judges, preventing the scorer from co-evolving with the policy. ARCO addresses all three axes with a single rubric model $\mu$ whose two branches instantiate $\rho$ and $g$ on a
shared backbone: the generation branch $\rho$ produces context-specific criteria \emph{per step}, the score branch $g$ predicts \emph{step-level} scores conditioned on those criteria, and $\mu$ co-evolves with the policy $\pi$ on on-policy data throughout training.

%% file: sections/method.tex
\section{ARCO}
\label{sec:method}
\begin{figure*}[t]
\centering
\includegraphics[
    width=\textwidth,
    trim=18pt 160pt 13pt 0pt,
    clip
]{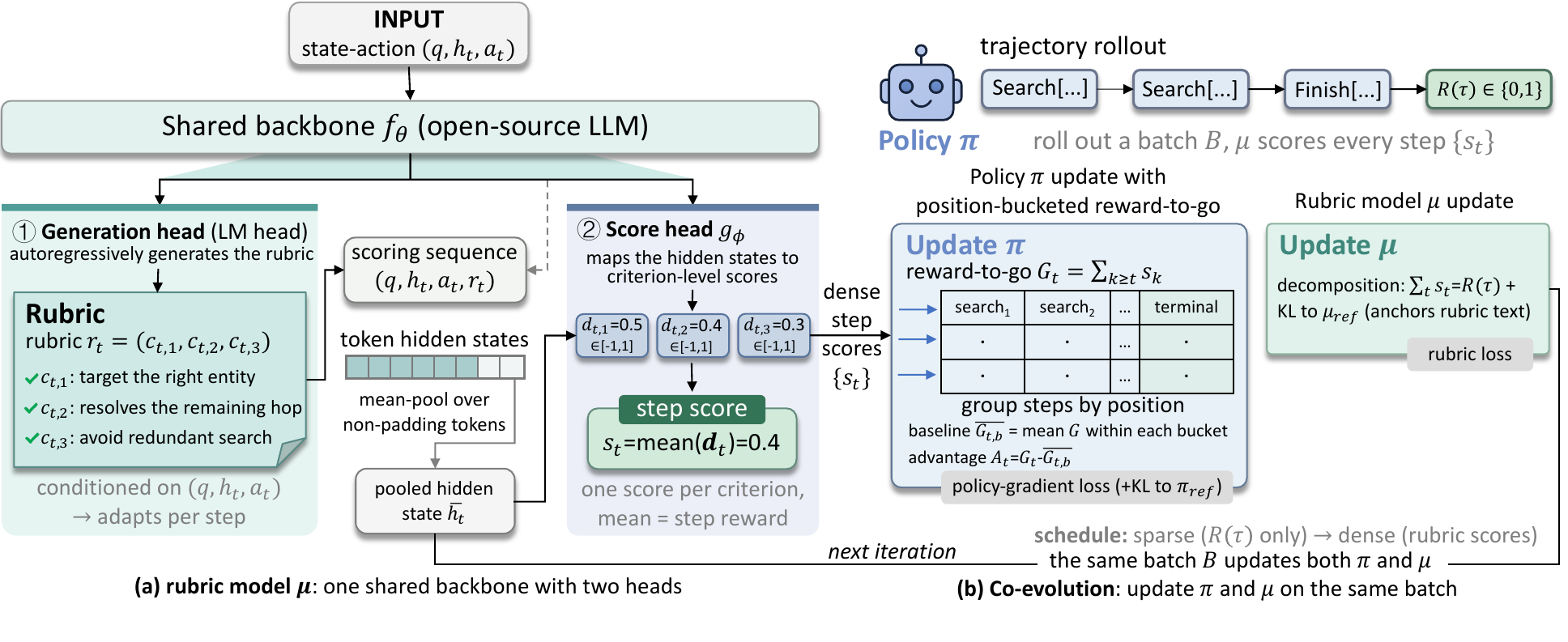}
\vspace{-20pt}
\caption{Overview of ARCO. \textbf{(a) Rubric model $\mu$}: from the state--action input at step $t$, a single open-source backbone with two heads produces a per-step rubric and score---a \emph{generation head} writes the rubric $r_t$ as a natural-language checklist, and a \emph{score head} scores each criterion and averages them into the step score $s_t$. \textbf{(b) Co-evolution}: on the \emph{same} batch, $\pi$ is updated from the reward-to-go of $\{s_t\}$ with a position-bucketed baseline (Eq.~\ref{eq:pi_rl}), and $\mu$ by a trajectory-decomposition loss with a KL that anchors the rubric text (Eq.~\ref{eq:mu_rl}); both models thus co-evolve on shared rollouts.}
\label{fig:training}
\vspace{-10pt}
\end{figure*}
We study an LLM-based agent that solves a multi-hop question $q\in\mathcal{Q}$ over multiple steps, following the notation of Section~\ref{sec:preliminary}: at step $t$ the policy $\pi$ selects an action $a_t$ from the prefix $h_t$, and the trajectory $\tau$ receives a binary outcome reward $R(\tau)\in\{0,1\}$ only at the end. ARCO turns this sparse signal into interpretable process supervision by attaching a natural-language rubric $r_t$ and a scalar score $s_t$ to the action at each step. Unlike pipelines that read rubrics with a frozen external judge, ARCO instantiates the generator $\rho$ and the scorer $g$ as two heads of a single trainable model $\mu$ that co-evolves with $\pi$. Figure~\ref{fig:training} illustrates our method and its co-evolution training procedure.

\subsection{Rubric Model}
\label{sec:method_model}
The rubric model $\mu$ follows two design principles. First, evaluation criteria should be \emph{generated} rather than fixed, because step quality is operation-dependent and its failure modes shift as the policy
improves: a \texttt{Search} may fail by wrong-entity targeting or redundancy, a \texttt{Finish} by missing support or wrong format, and a static checklist quickly goes stale. We therefore let a language model write a
fresh, context-specific rubric for each step, making the evaluator adaptive. Second, the rubric and the score should improve together rather than come from disconnected modules, so we place generation and scoring in a
\emph{single} model on a shared backbone, where the scoring gradient reshapes the representation that rubric generation also draws on. Concretely, $\mu$ adds two heads to a causal LM backbone $f_\theta$: a
\emph{generation head} that produces the rubric $r_t$, and a \emph{score head} $g_\phi$ that rates how well the action satisfies each criterion and outputs the step score $s_t$. In effect, $\mu$ writes a local
checklist for the current prefix--action pair and scores against it, exposing its latent dimensions as readable criteria.

We now specify each head in turn. 
The generation head acts as the rubric generator: given $(q,h_t,a_t)$, it generates an ordered rubric $r_t=(c_{t,1},\ldots,c_{t,K})\sim p_\theta(\cdot\mid q,h_t,a_t)$, a set of $K$ criteria that each name one aspect of action quality. Because the rubric is generated after
observing the prefix, it adapts across stages: early searches are judged by entity targeting and evidence coverage, later searches by whether they resolve the remaining hop, and finish actions by answer support and format. The score head acts as the evaluator: it reads the full scoring sequence
$(q,h_t,a_t,r_t)$ through the shared backbone, forms $\bar{\mathbf{h}}_t$ by attention-mask mean pooling over the non-padding hidden states, and maps it to $K$ criterion-level scores whose mean is the step reward,
\begin{equation}
\label{eq:step_score}
\mathbf{d}_t = g_\phi(\bar{\mathbf{h}}_t) \in [-1,1]^K,
\qquad
s_t = \frac{1}{K}\sum_{j=1}^{K} d_{t,j},
\end{equation}
where $d_{t,j}\in[-1,1]$ rates how well action $a_t$ satisfies criterion $c_{t,j}$, the vector $\mathbf{d}_t$ collects these $K$ scores, and their mean $s_t$ is the step-level reward that supervises the policy (Figure~\ref{fig:training}a). Sharing the backbone lets the two heads improve together: since $s_t$
regresses on $\bar{\mathbf{h}}_t=f_\theta(q,h_t,a_t,r_t)$, the scoring gradient updates both $g_\phi$ and the backbone $\theta$ that produces the rubric, so pressure to score accurately reshapes the representation generation depends on. This coupling is confined to \emph{optimization}: because $r_t$ is a
discrete sample, the gradient does not reach the rubric \emph{text}, which is instead anchored by warmup imitation and a KL term (\S\ref{sec:method_training}). Splitting $\mu$ into separate generator and scorer would sever this coupling and revert to the frozen-judge pipeline of existing rubric methods.

\subsection{Co-Evolutionary Training}
\label{sec:method_training}

We warm up both models by supervised fine-tuning (SFT) on rollouts from a closed-source LLM teacher in the same retrieval environment: $\pi$ imitates the actions of high-reward trajectories, while $\mu$ learns to
generate the annotated criteria and regress their scores, projected so that the per-step means satisfy the trajectory decomposition $\sum_t s_t = R(\tau)$ (Appendix~\ref{app:warmup}). The resulting checkpoints
initialize the evolving $\pi$ and $\mu$; frozen copies serve as the references $\pi_{\mathrm{ref}}$ and $\mu_{\mathrm{ref}}$ used by the KL terms below, and the same warmup is applied to all baselines for fairness.
The remainder of this section describes the on-policy co-evolution loop (Figure~\ref{fig:training}b), which updates $\pi$ and $\mu$ on the \emph{same} rollouts so that the evaluator tracks the policy's shifting
failure modes.

Each iteration proceeds in three steps: the current $\pi$ rolls out a batch $\mathcal{B}$ of trajectories, the current $\mu$ annotates every active step with a rubric and a score, and both models are updated on this
shared batch. Training follows a \emph{sparse-to-dense} schedule: the first few epochs optimize $\pi$ against the outcome reward $R(\tau)$ alone, giving $\mu$ time to catch up to fresh rollouts before its scores drive
the policy; the dense rubric scores take over thereafter.

The policy is updated by credit assignment on the dense scores. Within a trajectory, we use the reward-to-go $G_t=\sum_{k\ge t} s_k$, which propagates the consequences of an action forward to the steps it enables. A
global mean baseline is ill-suited here, since step scores vary systematically by position---early searches, later searches, and the terminal action occupy different ranges---so a single average would confound a weak
step with a merely late one. We therefore center each reward-to-go on a \emph{position-bucketed} baseline $\bar{G}_b$, the mean reward-to-go over all steps in the batch that share bucket $b$, where a step's bucket is
its search rank or \texttt{terminal} for the trajectory's last action (a successful \texttt{Finish} or a forced finish at budget exhaustion). The resulting advantage $A_t=\alpha\,(G_t-\bar{G}_{b(t)})$, with $\alpha$
scaling the dense reward, enters a KL-regularized policy-gradient objective,
\begin{equation}
\label{eq:pi_rl}
\mathcal{L}_{\pi}^{\mathrm{rl}}
= -\frac{1}{|\mathcal{B}|}
\sum_{\tau \in \mathcal{B}}
\sum_{t \in \mathcal{V}_{\tau}}
A_{t}\,
\log p_\pi(a_t \mid h_t)
+ \lambda_{\pi}\,\mathcal{L}_{\mathrm{KL}}^{\pi},
\end{equation}
where $\mathcal{V}_{\tau}$ is the set of valid policy steps in $\tau$, and $\lambda_\pi$ weights a token-level KL term $\mathcal{L}_{\mathrm{KL}}^{\pi}$ against the frozen reference $\pi_{\mathrm{ref}}$ that keeps the
policy from drifting away from coherent language.

The rubric model is updated on the same batch by a trajectory-decomposition loss regularized by a rubric-anchoring KL,
\begin{equation}
\label{eq:mu_rl}
\mathcal{L}_{\mu}^{\mathrm{rl}}
= \frac{1}{|\mathcal{B}|}
\sum_{\tau \in \mathcal{B}}
\Bigl(\textstyle\sum_{t} s_t - R(\tau)\Bigr)^{2}
+ \lambda_{\mu}\,
\mathbb{E}_{(h_t,a_t)}\!\Bigl[\textstyle\sum_{i}
\mathrm{KL}\bigl(p_\theta \,\Vert\, p_{\theta_{\mathrm{ref}}}\bigr)_{i}\Bigr].
\end{equation}
The decomposition term forces the step scores to sum to the observed outcome, learning per-step credit without any gold step labels, while $\lambda_\mu$ weights a per-token KL (summed over the rubric tokens, indexed
by $i$) that keeps generated rubrics close to the warmup language;
without it, $\mu$ could lower the decomposition loss by degenerating the rubric into meaningless tokens that happen to yield convenient scores, so $\mu_{\mathrm{ref}}$
anchors the wording while the score head absorbs the numerical fit.   Because $\pi$ and $\mu$ share rollouts, every policy improvement exposes new failure modes to the evaluator, whose updated scores in turn reshape the rewards that drive the next policy step, so the two co-evolve
  throughout training. Algorithm~\ref{alg:arco} in Appendix~\ref{app:algorithm} gives the full loop.

%% file: sections/experiments.tex
\section{Experiments}
\label{sec:experiments}

We evaluate ARCO on three multi-hop QA benchmarks across two open-source backbone families. We organize the analysis around five research questions:
\begin{itemize}[leftmargin=*,nosep]
\item \textbf{RQ1.} Does ARCO outperform existing outcome-, rubric-, and process-reward baselines?
\item \textbf{RQ2.} Is each of ARCO's design choices individually necessary?
\item \textbf{RQ3.} Do ARCO's rubrics actually point at the action they scored?
\item \textbf{RQ4.} How sensitive is ARCO to the number of rubric criteria $K$?
\item \textbf{RQ5.} How do rubric-model architecture and scale affect policy learning?
\end{itemize}

\subsection{Experimental Setup}

\paragraph{Datasets.}
We evaluate on three multi-hop QA benchmarks of increasing reasoning complexity: \textbf{HotpotQA}~\cite{yang2018hotpotqa} (2-hop), \textbf{2WikiMultiHopQA}~\cite{ho2020constructing} (multi-hop comparison/bridge), and \textbf{MuSiQue}~\cite{trivedi2022musique} (compositional, 2--4 hops). For each dataset we use $2{,}000$ training and $500$ evaluation examples and report Exact Match (EM) and token-level F1.

\paragraph{Agent environment.}
Following prior work~\cite{bo2024copper,tian2026haps,yao2023react,jin2025searchr1}, we adopt a unified multi-turn retrieval-augmented QA environment. At each step, the agent selects \texttt{Search[query]} or \texttt{Finish[answer]}. A \texttt{Search} returns the top-$1$ passage retrieved by a SimCSE retriever~\cite{gao2021simcse} over the supporting and distractor paragraphs supplied with each example, so retrieval is confined to the dataset's own context rather than an external corpus. The search budget is $3$ for HotpotQA and 2WikiMultiHopQA and $6$ for MuSiQue; the environment returns binary EM as $R(\tau)\in\{0,1\}$. Full interaction protocol, including forced-finish handling and rubric scoring, is given in Algorithm~\ref{alg:arco} of Appendix~\ref{app:algorithm}.

\paragraph{Implementation details.}
We instantiate ARCO with two same-scale open-source backbones, Qwen3-4B-Instruct-2507~\cite{qwen3} and Llama-3.2-3B-Instruct~\cite{llama3} (Qwen3-4B and Llama-3.2-3B for short); in each setting $\pi$ and $\mu$ share the base model and are adapted with LoRA~\cite{Hu2021LoRALA}, with the generation head
producing $K{=}3$ criteria per step. For the supervised warmup we annotate trajectories with two separate teachers: a weaker policy teacher (\texttt{gpt-4o-mini}) rolls out the trajectories, and a stronger rubric teacher (\texttt{gpt-5.4}) writes the per-step rubrics and criterion scores. We deliberately
pair the two: a near-perfect policy teacher would leave almost no faulty actions, and thus no negative examples for $\mu$ to learn to penalize, while the stronger rubric teacher keeps the supervision reliable. RL then follows the sparse-to-dense schedule of \S\ref{sec:method_training}. For each (method,
dataset, backbone) cell we report EM/F1 at the best post-transition dev-EM checkpoint; dense-stage wall-clock is in Appendix~\ref{app:runtime} and full hyperparameters in Appendix~\ref{app:warmup}. 

\paragraph{Baselines.}
We compare ARCO against a comprehensive set of baselines organized by reward granularity. Together these baselines span the three design axes of \S\ref{sec:preliminary}---rubric adaptation, scoring granularity, and judge model---so each comparison isolates one axis that ARCO changes. We first include a \textbf{Base Model} that uses zero-shot prompting without fine-tuning as a reference point, and then compare against two groups of trained baselines. \textbf{Group I (Outcome Reward Models, ORMs)}: \textbf{Search-R1}~\cite{jin2025searchr1} and \textbf{R1-Searcher}~\cite{song2025r1searcher} train with binary EM; three rubric-based ORMs score \emph{at the trajectory level} with a frozen closed-source judge but differ in rubric source: \textbf{RaR}~\cite{gunjal2025rubrics} pre-generates a query-specific rubric and keeps it fixed; \textbf{CARMO}~\cite{gupta2025carmo} generates context-aware criteria at scoring time conditioned on the query and candidate trajectory; \textbf{RLER}~\cite{shao2025dr} maintains an on-policy rubric buffer with frozen judge parameters. \textbf{Group II (Process Reward Models, PRMs)}: \textbf{AgentPRM}~\cite{xi2025agentprm} estimates step values via temporal-difference plus generalized-advantage-estimation (TD+GAE) and is frozen during RL. \textbf{ARCO} (ours) generates per-step rubric criteria and predicts rubric-conditioned step scores with a same-scale open-source $\mu$ that co-evolves with $\pi$.

\subsection{Overall Performance (RQ1)}
\label{sec:main_results}

\begin{table*}[t]
\centering
 \caption{Main results on three multi-hop QA benchmarks. All values are reported as percentages, with the ``$\%$'' symbol omitted for brevity. Rubric-based methods are highlighted in light blue. The best result in each column is shown in \textbf{bold}, and the second-best is \underline{underlined}.}
\vspace{-8pt}
\label{tab:main_results}
\setlength{\tabcolsep}{3.5pt}
\renewcommand{\arraystretch}{1.15}
\resizebox{\textwidth}{!}{%
\begin{tabular}{@{}l cc cc cc cc cc cc@{}}
\toprule
& \multicolumn{4}{c}{\textbf{HotpotQA}}
& \multicolumn{4}{c}{\textbf{2WikiMultiHopQA}}
& \multicolumn{4}{c}{\textbf{MuSiQue}} \\
\cmidrule(lr){2-5} \cmidrule(lr){6-9} \cmidrule(lr){10-13}
& \multicolumn{2}{c}{Qwen3-4B} & \multicolumn{2}{c}{Llama-3.2-3B}
& \multicolumn{2}{c}{Qwen3-4B} & \multicolumn{2}{c}{Llama-3.2-3B}
& \multicolumn{2}{c}{Qwen3-4B} & \multicolumn{2}{c}{Llama-3.2-3B} \\
\cmidrule(lr){2-3} \cmidrule(lr){4-5}
\cmidrule(lr){6-7} \cmidrule(lr){8-9}
\cmidrule(lr){10-11} \cmidrule(lr){12-13}
\textbf{Method}
& EM & F1 & EM & F1
& EM & F1 & EM & F1
& EM & F1 & EM & F1 \\
\midrule
Base Model
& 34.40 & 45.93 & 18.40 & 29.48
& 47.80 & 53.95 & 26.00 & 32.54
& 19.40 & 28.05 & 9.40 & 15.12 \\
\midrule
\multicolumn{13}{l}{\emph{Group I: Outcome Reward Models (ORM)}} \\
Search-R1~\cite{jin2025searchr1}
& 39.40 & 51.67 & 34.60 & \underline{45.54}
& \underline{63.40} & \underline{67.95} & \underline{58.20} & \underline{62.17}
& 24.80 & 33.66 & \underline{18.80} & \underline{26.99} \\
R1-Searcher~\cite{song2025r1searcher}
& \underline{41.00} & \underline{53.42} & \underline{35.00} & 44.48
& 62.00 & 67.19 & 57.80 & 61.48
& 26.00 & \underline{36.02} & 17.40 & 26.41 \\
\rowcolor{retroBlue!12}
RaR~\cite{gunjal2025rubrics}
& 38.00 & 50.75 & 29.60 & 40.83
& 61.40 & 67.06 & 53.80 & 57.88
& \underline{26.60} & \textbf{36.46} & 16.00 & 25.26 \\
\rowcolor{retroBlue!12}
CARMO~\cite{gupta2025carmo}
& 34.80 & 48.52 & 30.20 & 41.74
& 59.40 & 65.77 & 51.60 & 58.15
& 22.20 & 32.26 & 13.40 & 21.49 \\
\rowcolor{retroBlue!12}
RLER~\cite{shao2025dr}
& 32.20 & 44.40 & 28.80 & 41.31
& 61.00 & 66.06 & 50.00 & 55.00
& 24.40 & 33.35 & 14.60 & 23.02 \\
\midrule
\multicolumn{13}{l}{\emph{Group II: Process Reward Models (PRM)}} \\
AgentPRM~\cite{xi2025agentprm}
& 38.60 & 51.49 & 29.40 & 37.29
& 62.00 & 65.32 & 49.60 & 54.63
& 25.80 & 34.81 & 16.60 & 25.83 \\
\rowcolor{retroBlue!12}
\textbf{ARCO} (ours)
& \textbf{42.80} & \textbf{55.07} & \textbf{36.40} & \textbf{45.58}
& \textbf{65.20} & \textbf{69.38} & \textbf{59.60} & \textbf{63.67}
& \textbf{27.40} & 35.55 & \textbf{21.60} & \textbf{29.70} \\
\bottomrule
\end{tabular}%
}
% \vspace{-10pt}
\end{table*}

Table~\ref{tab:main_results} compares ARCO with outcome-level RL baselines, rubric-based ORMs, and a process reward model under the same protocol. ARCO achieves the best EM in all six (dataset, backbone) cells and the best F1 in five of six. The advantage is thus broad across backbones and metrics rather than concentrated in one setting.
Among the outcome-level groups, the binary-reward ORMs Search-R1 and R1-Searcher are strong and often outperform the frozen or trajectory-level rubric ORMs. ARCO improves over both, with a larger EM margin over the rubric ORMs (RaR, CARMO, RLER) than over the binary-reward ones: adding a static or
frozen-judge rubric on top of outcome-level training is not enough, since these judges still collapse the whole rollout into a single score. This supports ARCO's central design choice---the rubric and scorer must co-evolve with the policy at the \emph{step level}, not serve as a fixed outcome-level annotation layer.
ARCO also outperforms the step-level PRM AgentPRM by $1.6$--$10.0$ EM across all six cells while preserving natural-language interpretability, showing that process-level credit assignment need not rely on opaque scalars. The gain holds on MuSiQue, the most compositional benchmark ($2$--$4$ hops), where ARCO
leads EM on both backbones ($+0.80$ Qwen, $+2.80$ Llama over the strongest baseline).

\subsection{Ablation Study (RQ2)}
\label{sec:ablation}

\begin{table*}[t]
\centering
\caption{Ablation study. \textbf{w/o Rubric}\,=\,remove rubric text generation and predict scores directly from trajectory context; \textbf{w/o Adaptive}\,=\,replace per-step generated rubric with three fixed dataset-level criteria reused across all steps; \textbf{w/o Prefix}\,=\,let $\mu$ generate the rubric per step but condition only on the question and current action, with the trajectory prefix $h_t$ removed from $\mu$'s prompt.}
\vspace{-10pt}
\label{tab:ablation}
\setlength{\tabcolsep}{3.5pt}
\renewcommand{\arraystretch}{1.10}
\resizebox{\textwidth}{!}{%
\begin{tabular}{@{}l cc cc cc cc cc cc@{}}
\toprule
& \multicolumn{4}{c}{\textbf{HotpotQA}}
& \multicolumn{4}{c}{\textbf{2WikiMultiHopQA}}
& \multicolumn{4}{c}{\textbf{MuSiQue}} \\
\cmidrule(lr){2-5} \cmidrule(lr){6-9} \cmidrule(lr){10-13}
& \multicolumn{2}{c}{Qwen3-4B} & \multicolumn{2}{c}{Llama-3.2-3B}
& \multicolumn{2}{c}{Qwen3-4B} & \multicolumn{2}{c}{Llama-3.2-3B}
& \multicolumn{2}{c}{Qwen3-4B} & \multicolumn{2}{c}{Llama-3.2-3B} \\
\cmidrule(lr){2-3} \cmidrule(lr){4-5}
\cmidrule(lr){6-7} \cmidrule(lr){8-9}
\cmidrule(lr){10-11} \cmidrule(lr){12-13}
\textbf{Variant}
& EM & F1 & EM & F1
& EM & F1 & EM & F1
& EM & F1 & EM & F1 \\
\midrule
SFT Only ($\pi_0$)
& 35.40 & 47.04 & 28.00 & 38.03
& 60.80 & 66.07 & 51.00 & 54.49
& 23.20 & 31.79 & 12.60 & 20.72 \\
\midrule
w/o Rubric
& 39.80 & 52.67 & 35.20 & 45.16
& 62.40 & 67.55 & 59.00 & 62.51
& 26.00 & 34.58 & 20.20 & 28.16 \\
w/o Adaptive
& 38.00 & 50.41 & 35.00 & 44.56
& 63.80 & 68.11 & 59.20 & 62.67
& \underline{26.80} & \textbf{35.61} & 19.80 & 29.11 \\
w/o Prefix
& \underline{41.60} & \underline{54.28} & \underline{35.40} & \underline{45.43}
& \underline{64.60} & \underline{68.64} & \underline{59.40} & \underline{63.37}
& 25.60 & 34.31 & \underline{20.60} & \textbf{29.81} \\
\midrule
\textbf{Ours (Full)}
& \textbf{42.80} & \textbf{55.07} & \textbf{36.40} & \textbf{45.58}
& \textbf{65.20} & \textbf{69.38} & \textbf{59.60} & \textbf{63.67}
& \textbf{27.40} & \underline{35.55} & \textbf{21.60} & \underline{29.70} \\
\bottomrule
\end{tabular}%
}
\vspace{-10pt}
\end{table*}

\begin{figure}[t]
\centering
\includegraphics[
    width=\textwidth,
    trim=20pt 100pt 20pt 102pt,
    clip
]{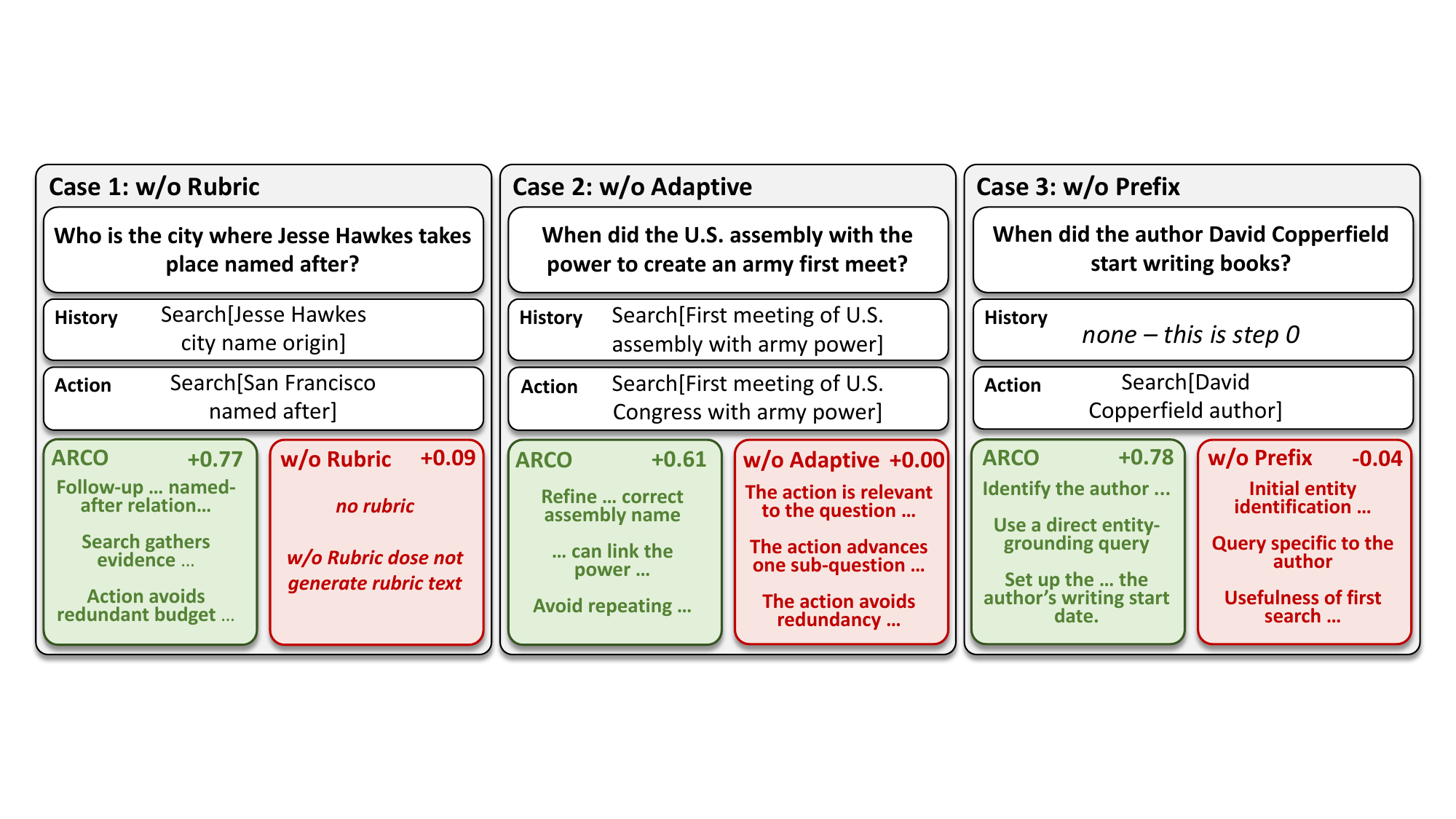}
\vspace{-15pt}
\caption{Three single-step score comparisons on MuSiQue dev. In each panel, ARCO and one ablation see the same $(h_t, a_t)$; only the evaluator differs. ARCO scores all three steps well, while each variant fails for a different reason and produces a mismatched score.}
\label{fig:ablation_cases}
% \vspace{-20pt}
\end{figure}

We ablate three core components of ARCO, with results in Table~\ref{tab:ablation}. \textbf{w/o Rubric} removes generated rubric text and predicts step scores directly from a mean-pooled trajectory-action representation; \textbf{w/o Adaptive} uses a fixed set of dataset-level criteria for every step instead
of generating criteria per action; and \textbf{w/o Prefix} still generates per-step rubrics but conditions $\mu$ only on the question and current action, excluding the trajectory prefix. All three contribute. \textbf{w/o Rubric} drops EM by $3.0\%$--$3.5\%$ on HotpotQA, confirming that rubric text provides
useful semantic structure rather than a decorative explanation. \textbf{w/o Adaptive} matches full ARCO on 2WikiMultiHopQA but lags by $4.80\%$ EM on HotpotQA/Qwen ($38.00$ vs.\ $42.80$), showing that fixed criteria cannot always track the action-specific failure modes seen during training. \textbf{w/o
Prefix} is the strongest ablation, slightly improving a few F1 cells (e.g., MuSiQue/Llama F1 $29.81$ vs.\ $29.70$) but losing EM in every cell, suggesting prefix-conditioned rubrics mainly improve correctness. Figure~\ref{fig:ablation_cases} shows three qualitative examples from MuSiQue dev: since each
ablation trains a separate policy, we select matched steps where the policy state, step index, and action coincide, so both evaluators see the same $(h_t,a_t)$ and the score difference is attributable to the evaluator design rather than different rollout trajectories.

\subsection{Rubric Quality: Do Rubrics Identify the Action They Scored? (RQ3)}
\label{sec:rq3_rubric_quality}
\input{sections/rq3_quality_table.tex}
A rubric that is genuinely about a specific action ought to make that action recoverable. Since no existing metric measures whether a generated rubric is bound to the step it scored, we propose a four-way step-binding diagnostic built from ARCO's peak-dev-EM rollouts: the gold option is the action the rubric
was generated for, and the three distractors are other unique actions from the same trajectory, so the judge must rely on the rubric text to identify it. We report two metrics, both judged by \texttt{gpt-4o-mini}: \textbf{Step-Binding Accuracy} (Bind, $\%$), the accuracy of selecting the gold action from
the rubric and four candidates alone (chance $25\%$), and \textbf{Step-Specificity} (Spec, $1$--$5$), rating how specifically the rubric evaluates the gold action given the question and prefix; both are reported overall and split by \texttt{Search} and \texttt{Finish} steps. Table~\ref{tab:rq3_quality} shows
ARCO's rubrics are clearly action-bound: Bind ranges from $42.46\%$ to $54.80\%$, roughly $1.7$--$2.2\times$ chance, while Spec stays near $4.0$. HotpotQA is strongest, with Qwen reaching $54.80\%$ Bind and both backbones around or above $50\%$; 2Wiki and MuSiQue are harder, since adjacent hops can produce
semantically similar \texttt{Search} actions for which a specific rubric stays plausible for a nearby distractor. Counterintuitively, \texttt{Finish} Bind can fall below \texttt{Search} Bind in some settings, though final-answer decisions should in principle be easier to identify.
Appendix~\ref{app:case_figures} provides a qualitative analysis with successful and failed \texttt{Search}/\texttt{Finish} examples, suggesting directions for more action-discriminative rubrics; Appendix~\ref{app:step_signal} further analyzes how diagnostic the step \emph{scores} are of final success.

\subsection{Rubric Width: How Many Criteria Per Step? (RQ4)}
\label{sec:rq4_rubric_K}

\begin{figure}[t]
\centering
\includegraphics[width=\linewidth]{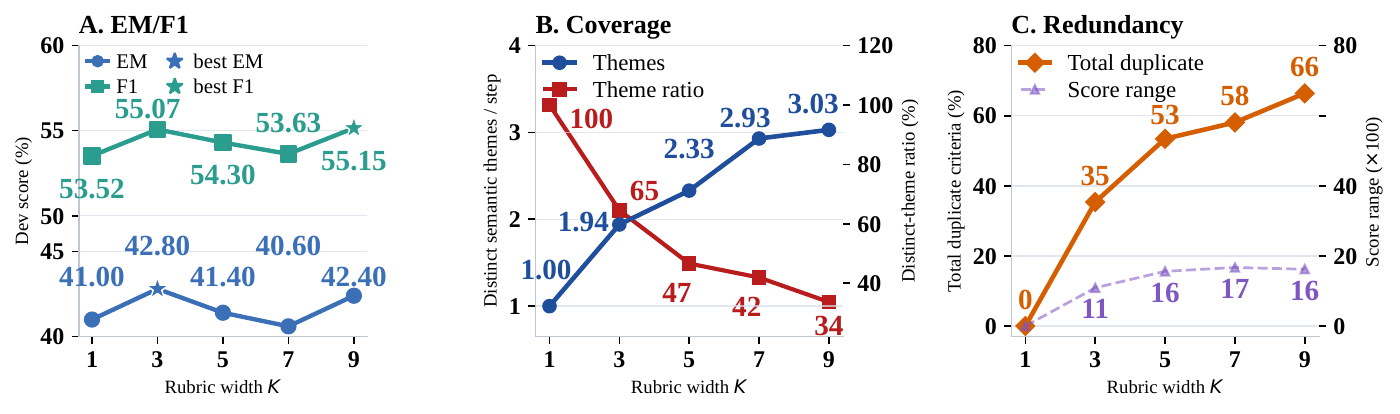}
% \vspace{-20pt}
\caption{RQ4 $K$-sensitivity diagnostics on HotpotQA / Qwen for the sweep on fixed policy-warmup trajectories ($K\in\{1,3,5,7,9\}$). (A) Dev EM/F1 at each best-dev-EM checkpoint. (B) Distinct semantic themes per step (blue, left) and distinct-theme ratio $\#\text{themes}/K$ (red, right). (C) Step-level total semantic duplicate rate (orange, left) and score range (purple, right; max--min scores within same-step duplicate semantic clusters, scaled by $100$), both averaged over dev steps.}
\label{fig:rq4_k_sensitivity}
\vspace{-15pt}
\end{figure}

The rubric turns ARCO's process reward into a readable step-level signal, and the most direct knob over its shape is the width $K$---how many criteria $\mu$ verbalizes per action. Our main experiments use $K{=}3$;
here we ask how this width affects downstream learning. To isolate $K$, we reuse the HotpotQA policy-warmup trajectories and re-run \emph{only} the rubric annotation under $K\in\{1,3,5,7,9\}$, repeating the same $\mu$ SFT and RL recipe and selecting the best-dev-EM checkpoint. We diagnose rubric content by embedding each criterion and running K-Means within every dev step, computing distinct themes, duplicate rate, and within-cluster score range over the resulting clusters. Increasing $K$ does not monotonically improve ARCO (Figure~\ref{fig:rq4_k_sensitivity}A): EM peaks at $42.80$ for $K{=}3$, drops at $K{=}5$/$7$, and only partially recovers at $K{=}9$. Distinct themes saturate quickly: the theme ratio roughly halves between $K{=}3$ and $K{=}9$, so nine criteria still cover about three themes on average (Figure~\ref{fig:rq4_k_sensitivity}B), and duplicates dominate at large $K$ (Figure~\ref{fig:rq4_k_sensitivity}C). These fall into three patterns---\emph{exact duplicates} (byte-identical criteria), \emph{semantic duplicates} (different wordings scoring the same theme), and \emph{diverse} rubrics (criteria spanning distinct themes); Appendix~\ref{app:rubric_redundancy} shows a concrete step of each (Figure~\ref{fig:rq4_duplicate_cases}). Byte-identical duplicates are rare; the redundancy at large $K$ is overwhelmingly semantic, while the diverse pattern we would ideally want becomes the minority as $K$ grows. The within-cluster score range is nonzero but modest on $[-1,1]$, so this semantic redundancy is not contradictory noise but adaptive reweighting, with $\mu$ emphasizing important themes through paraphrase rather than diversity. 

\subsection{Backbone Asymmetry: Rubric Architecture and Scale (RQ5)}
\label{sec:rq5_pi_mu_asymmetry}

\begin{wraptable}{r}{0.45\textwidth}
\vspace{-10pt}
\centering

\caption{RQ5 results on HotpotQA. Top: $\mu$-size sweep with $\pi$ fixed to Qwen3-4B. Bottom: cross-family pairs at comparable scale (Qwen3-4B vs.\ Llama-3.2-3B).}
\vspace{-5pt}
\label{tab:rq5_asymmetry}
\setlength{\tabcolsep}{3.0pt}
\renewcommand{\arraystretch}{1.02}
\footnotesize
\begin{tabular*}{\linewidth}{@{\extracolsep{\fill}}llcc@{}}
\toprule
Policy & Rubric & EM & F1 \\
\midrule
Qwen3-4B & Qwen3-0.6B & \textbf{42.80} & 54.69 \\
Qwen3-4B & Qwen3-1.7B & 42.40 & 54.54 \\
Qwen3-4B & Qwen3-4B & \textbf{42.80} & \textbf{55.07} \\
Qwen3-4B & Qwen3-8B & 42.00 & 54.32 \\
\midrule
Qwen3-4B & Qwen3-4B & \textbf{42.80} & \textbf{55.07} \\
Qwen3-4B & Llama-3.2-3B & 41.40 & 54.44 \\
Llama-3.2-3B & Qwen3-4B & 37.20 & 47.60 \\
Llama-3.2-3B & Llama-3.2-3B & 36.40 & 45.58 \\
\bottomrule
\end{tabular*}
\vspace{-15pt}
\end{wraptable}

Throughout the main table $\pi$ and $\mu$ share the same base model, entangling the rubric's contribution with the policy backbone. RQ5 disentangles them and asks whether final policy quality is controlled by the rubric model's size or family: we fix the policy to Qwen3-4B and vary only the Qwen3-family $\mu$ size, then keep the scale comparable and swap the backbone family in $(\pi,\mu)$ pairings on HotpotQA.

The size sweep is counterintuitive. Policy quality is nearly flat rather than monotonic: the $0.6$B and $4$B rubrics tie at $42.80$ EM, and the $8$B rubric does not improve further. Rubric modeling is not capacity-limited once $\mu$ can generate action-specific criteria; a smaller $\mu$ may even be easier to keep synchronized with a changing policy, providing rewards closer to the current rollout distribution.
The cross-family comparison probes whether evaluator-family quality matters beyond backbone sharing at comparable scale. For the Qwen policy, swapping the Qwen rubric for Llama lowers EM from $42.80$ to $41.40$; for the Llama policy, swapping Llama for Qwen as $\mu$ improves EM from $36.40$ to $37.20$ and F1 from $45.58$ to $47.60$. ARCO thus does not require $\pi$ and $\mu$ to share a backbone, but a stronger evaluator family transfers useful scoring signals across policy backbones. Overall, RQ5 suggests that ARCO benefits more from evaluator quality and co-evolutionary responsiveness than from scaling up $\mu$.

%% file: sections/rq3_quality_table.tex
% Auto-generated by scripts/rqs/rq3_rubric_analysis/make_rq3_table.py
\begin{table*}[t]
\centering
\caption{RQ3 step-level rubric quality. \textbf{Bind}: rubric-only four-way matching accuracy ($\%$, chance $25.0$). \textbf{Spec}: step-specificity ($1$--$5$). $\pm$ are 95\% CI half-widths.}
\vspace{-5pt}
\label{tab:rq3_quality}
\newcommand{\rqci}[2]{#1\smash{\raisebox{-0.45ex}{\scriptsize\,$\pm #2$}}}
\small
\setlength{\tabcolsep}{3.5pt}
\renewcommand{\arraystretch}{1.10}
\begin{tabular*}{\textwidth}{@{\extracolsep{\fill}}ll*{6}{l}@{}}
\toprule
& & \multicolumn{3}{c}{\textbf{Bind} ($\%$, $\uparrow$)} & \multicolumn{3}{c}{\textbf{Spec} ($1$--$5$, $\uparrow$)} \\
\cmidrule(lr){3-5} \cmidrule(lr){6-8}
Dataset & Backbone & ALL & Search & Finish & ALL & Search & Finish \\
\midrule
  \multirow{2}{*}{HotpotQA} & Qwen3-4B & \rqci{\textbf{54.80}}{4.34} & \rqci{\textbf{50.33}}{5.62} & \rqci{\textbf{61.50}}{6.68} & \rqci{4.10}{0.04} & \rqci{\underline{4.05}}{0.02} & \rqci{\underline{4.17}}{0.09} \\
   & Llama-3.2-3B & \rqci{\underline{50.60}}{4.36} & \rqci{43.67}{5.58} & \rqci{\underline{61.00}}{6.70} & \rqci{\underline{4.11}}{0.04} & \rqci{4.02}{0.02} & \rqci{\textbf{4.25}}{0.08} \\
\midrule
  \multirow{2}{*}{2Wiki} & Qwen3-4B & \rqci{48.83}{3.99} & \rqci{46.31}{4.60} & \rqci{56.21}{7.76} & \rqci{4.04}{0.03} & \rqci{\underline{4.05}}{0.02} & \rqci{4.01}{0.09} \\
   & Llama-3.2-3B & \rqci{42.46}{6.05} & \rqci{36.59}{7.30} & \rqci{53.41}{10.21} & \rqci{\textbf{4.13}}{0.05} & \rqci{\textbf{4.11}}{0.06} & \rqci{\underline{4.17}}{0.10} \\
\midrule
  \multirow{2}{*}{MuSiQue} & Qwen3-4B & \rqci{44.76}{6.14} & \rqci{43.92}{7.90} & \rqci{46.00}{9.59} & \rqci{4.05}{0.04} & \rqci{4.01}{0.06} & \rqci{4.12}{0.06} \\
   & Llama-3.2-3B & \rqci{43.31}{6.05} & \rqci{\underline{47.95}}{8.00} & \rqci{37.04}{8.96} & \rqci{3.99}{0.05} & \rqci{3.97}{0.05} & \rqci{4.02}{0.10} \\
\bottomrule
\end{tabular*}
\vspace{-10pt}
\end{table*}

%% file: sections/related_work.tex
\section{Related Work}
\label{sec:related}

\paragraph{Reward modeling for multi-step LLM agents.}
RL for multi-step LLM agents typically uses scalar rewards from one of two families. \emph{Outcome reward models} (ORMs) inherit trajectory-level feedback from reinforcement learning from human feedback (RLHF) and direct preference optimization (DPO)~\cite{ouyang2022training,rafailov2023direct} and, in agent settings, train end-to-end with binary task success as the only signal~\cite{jin2025searchr1,song2025r1searcher,chen2026learning,zhao2025r,zeng2025reinforcing}. Such signals are too coarse for credit assignment across steps and are prone to reward hacking~\cite{gao2023scaling}. \emph{Process reward models} (PRMs) score every step: starting from math step verifiers~\cite{lightman2023let}, recent agent variants estimate step values via TD/GAE~\cite{xi2025agentprm,choudhury2025agentprm} or via tree-search, step-wise PPO, and implicit step supervision~\cite{xiong2024ipr,wang2025stepsearch,wang2025rro,liu2025agentic}. These methods address granularity but produce opaque scalars that do not explain why a step is good or bad.

\paragraph{Rubric-based rewards and joint policy--reward learning.}
To improve interpretability, \emph{rubric-based rewards} ground evaluation in natural-language criteria. RaR~\cite{gunjal2025rubrics} freezes a query-specific rubric before RL; CARMO~\cite{gupta2025carmo} generates rubrics at scoring time but freezes the generator; OpenRubrics~\cite{liu2025openrubrics}, Rubric Anchors~\cite{huang2025reinforcementlearningrubricanchors}, Chasing-the-Tail~\cite{zhang2025chasing}, and Online Rubrics Elicitation~\cite{rezaei2025online} focus on synthesizing or eliciting rubrics; RLER~\cite{shao2025dr} evolves a rubric buffer during training; RubricARM~\cite{xu2026rubricarm} adds alternating training. All score \emph{at the trajectory level} with a \emph{frozen closed-source judge}. A separate thread \emph{jointly trains policy and reward model}~\cite{hong2025cooper,shi2025mutual,huang2025efficient,wang2026co,wang2026rlanything}, but uses opaque scalars or no step-level supervision. \textbf{ARCO} unifies these axes: it generates \emph{step-level} natural-language rubrics with a same-scale open-source model, predicts \emph{step-level} scores, and---through trajectory decomposition---co-evolves rubric content and scoring with the policy at the \emph{parameter level}, with no external judge.

%% file: sections/conclusion.tex
\section{Limitations}
\label{sec:limitations}
ARCO trains the rubric model with lightweight objectives, and its process signal is fairly single-form: a per-step scalar tied to the outcome only through a simple sum constraint. Richer or more structured supervision could yield sharper per-step credit. The pace of $\pi$--$\mu$ co-evolution is likewise fixed by preset learning rates and regularization weights, leaving adaptive update ratios and drift-aware scheduling to future work. Finally, training both $\pi$ and $\mu$ and invoking $\mu$ at every active step makes ARCO more expensive than single-model RL, an overhead that grows with horizon length; distillation, rubric caching, or asynchronous $\mu$ updates could reduce it. These limitations point in the same direction: tighter, more adaptive coupling between the policy and an interpretable evaluator.

\section{Conclusion}
We presented ARCO, an adaptive rubric co-evolution framework that gives multi-step language agents interpretable step-level process rewards. A single rubric model couples a generation head and a score head on a shared backbone to write per-step criteria and score actions under them, while trajectory decomposition jointly optimizes the evaluator and policy from terminal outcomes. Because rubric generation, scoring, and policy improvement update the same backbone on the same on-policy rollouts, ARCO turns sparse outcome rewards into a dense, readable, step-aware signal with no external judge. Across multi-hop QA benchmarks and backbones, ARCO improves over outcome- and process-reward baselines, and its rubrics are step-specific, robust to design choices, and useful for diagnosing agent behavior. We hope ARCO offers a foundation for adaptive, interpretable, process-aware reward modeling for language agents.

%% file: sections/ethics_repro.tex
\section*{Ethics Statement}
This work uses publicly released benchmarks (HotpotQA, 2WikiMultiHopQA, MuSiQue) and publicly available open-source models (Qwen3-4B-Instruct-2507 and Llama-3.2-3B-Instruct), all employed in accordance with their respective licenses. The studies do not involve human subjects, personally identifiable information, or sensitive content. We use GPT API only as an automated annotator for warmup trajectories and step-level rubrics; all generated text is consumed inside our training pipeline and is not redistributed. ARCO targets the general problem of interpretable process-reward learning for multi-step language agents and does not enable a clear dual-use beyond what is already possible with existing rubric-based or process-reward methods. 

\section*{Reproducibility Statement}
We have made an effort to make ARCO reproducible. The full method is specified in Section~\ref{sec:method}, with the formal training loop in Algorithm~\ref{alg:arco} (Appendix~\ref{app:algorithm}). The agent environment, datasets, evaluation protocol, and best-checkpoint selection rule are described in Section~\ref{sec:experiments}; warmup trajectory collection, rubric-annotation projection, and the dual SFT objective are detailed in Appendix~\ref{app:warmup}; dense-stage wall-clock costs for every (method, dataset, backbone) cell are reported in Appendix~\ref{app:runtime}; and the policy, step-level rubric, and trajectory-level rubric prompts, together with the RQ3 judge prompts, are listed verbatim in Appendix~\ref{app:prompts}. Source code, training scripts, configuration files, and the warmup data used in the main-table HotpotQA / Qwen run are released at \href{https://github.com/zihangtian/ARCO}{https://github.com/zihangtian/ARCO}.

%% file: sections/appendix_warmup.tex
\section{Training Details}
\label{app:warmup}

This appendix details ARCO's two training stages: the supervised warmup that initializes $\pi$ and $\mu$, and the full on-policy co-evolution loop.

\subsection{Warmup Stage}

ARCO uses a supervised warmup before on-policy co-evolution so that both the policy and the rubric model start from a working initialization. We collect warmup records with two separate teachers: a weaker policy teacher (\texttt{gpt-4o-mini}) rolls out trajectories on the training questions in the same retrieval environment used for RL, and a stronger rubric teacher (\texttt{gpt-5.4}) annotates each complete trajectory with per-step criteria and scores. We deliberately pair a weaker policy teacher with a stronger rubric teacher: a near-perfect policy teacher would produce almost no faulty actions, leaving the rubric model with no negative examples to learn to penalize, while the stronger rubric teacher keeps the annotations reliable. For each question we sample multiple trajectories and log every action, observation, parse status, search budget, and terminal reward. HotpotQA and 2Wiki use a search horizon of three steps, MuSiQue uses six; the terminal reward is the task EM reward used by the main experiments.

\paragraph{Policy warmup.}
The policy SFT dataset is built only from the correct trajectories (those that end with EM${=}1$), converted into one supervised example per step. The input contains the question, the trajectory prefix before the current action, and the remaining search budget; when the budget is exhausted and the dataset supports forced finishing, the prompt switches to a finish-only template. The target is the recorded action string, e.g., \texttt{Search[query]} or \texttt{Finish[answer]}, which initializes $\pi$ to imitate successful tool-use traces rather than all sampled behavior.

\paragraph{Rubric warmup.}
The rubric annotator sees the full sampled trajectory and returns, for each step, $K$ natural-language criteria and $K$ criterion scores. Scores are parsed, clipped to $[-1,1]$, padded or truncated to width $K$, and projected so that the summed step-score means satisfy the trajectory-level decomposition constraint,
\begin{equation}
\label{eq:warmup_projection}
\sum_{t}\frac{1}{K}\sum_{j=1}^{K}\hat{s}_{t,j}=R(\tau),
\end{equation}
where $\hat{s}_{t,j}$ is the projected score of criterion $j$ at step $t$ and the vector $\hat{\mathbf{s}}_t=(\hat{s}_{t,1},\ldots,\hat{s}_{t,K})$ collects them for Eq.~\ref{eq:sft_mu}. The projection preserves relative score preferences while making the warmup data compatible with the later decomposition loss. The trajectory-level prompt is shown in Appendix~\ref{app:prompts}. Before $\mu$ SFT, records are filtered by the same constraint with a small tolerance, and steps with empty criteria are dropped.

The rubric model $\mu$ is then trained with a dual objective: the generation head learns to generate the criterion JSON for each state-action pair, while the score head regresses the annotated criterion scores from the mean-pooled hidden representation of the full scoring sequence. The SFT loss is
\begin{equation}
\label{eq:sft_mu}
\mathcal{L}_{\mu}^{\text{sft}}
=
\underbrace{-\sum_{t}\log p_\theta(\tilde{r}_t\mid q,h_t,a_t)}_{\text{rubric text loss}}
+\lambda_{\mathrm{sft}}
\underbrace{\sum_{t}\lVert \mathbf{d}_t-\hat{\mathbf{s}}_t\rVert_2^2}_{\text{criterion-score loss}},
\end{equation}
where $\tilde{r}_t$ is the annotated criterion list, $\hat{\mathbf{s}}_t\in[-1,1]^K$ the projected score vector, $\mathbf{d}_t$ the score-head prediction, and $\lambda_{\mathrm{sft}}$ balances the two terms (distinct from the RL-stage KL weights $\lambda_\pi,\lambda_\mu$ of \S\ref{sec:method_training}). The default $K{=}3$; ablations reuse the same pipeline but either remove the rubric text target or replace it with a fixed rubric.

After warmup, the selected policy checkpoint initializes the evolving $\pi$ and the frozen reference $\pi_{\mathrm{ref}}$, and the selected rubric checkpoint initializes the evolving $\mu$ and the frozen reference $\mu_{\mathrm{ref}}$; both references anchor the KL terms during RL.

\subsection{Full Training Algorithm}
\label{app:algorithm}

Algorithm~\ref{alg:arco} details the full ARCO training loop, covering rollout, rubric generation and scoring, and the co-evolution updates of $\pi$ and $\mu$.

\begin{algorithm}[h]
\caption{ARCO: Adaptive Rubric Co-Evolution}
\label{alg:arco}
\begin{algorithmic}[1]
\Require Policy $\pi$ (SFT-initialized), rubric model $\mu$ (SFT-initialized), training examples $\mathcal{E}$, retriever $\mathcal{R}$, max steps $T$, dense transition epoch $\eta$
\Statex
\For{epoch $= 1, \ldots, E$}
\State $\textit{use\_dense} \gets (\text{epoch} \ge \eta)$ \Comment{Sparse-to-dense schedule}
\For{batch $\mathcal{B} \subset \mathcal{E}$}
  \Statex \hspace{1.5em} \textit{// Phase 1: Rollout}
  \For{each example $q \in \mathcal{B}$}
    \State $h_0 \gets q$
    \For{$t = 1, \ldots, T$}
      \State $a_t \sim \pi(\cdot \mid h_t)$ \Comment{Sample action: \texttt{Search[query]} or \texttt{Finish[ans]}}
      \If{$a_t = \texttt{Search[query]}$}
        \State $o_t \gets \mathcal{R}(\text{query}, q.\text{paragraphs})$ \Comment{Dense retrieval over example paragraphs}
      \EndIf
      \State $h_{t+1} \gets (h_t, a_t, o_t)$
      \Statex \hspace{3em} \textit{// Phase 2: Rubric generation \& scoring}
      \State $r_t \sim p_\theta(\cdot \mid q, h_t, a_t)$ \Comment{$\mu$ generates $K{=}3$ criteria}
      \State $\bar{\mathbf{h}}_t \gets \mathrm{MeanPool}(\text{hidden}(q, h_t, a_t, r_t))$
      \State $\mathbf{d}_t \gets g_\phi(\bar{\mathbf{h}}_t)$; \quad $s_t \gets \text{mean}(\mathbf{d}_t)$ \Comment{Criterion scores and step score; Eq.~\ref{eq:step_score}}
      \If{$a_t = \texttt{Finish[\ldots]}$} \textbf{break} \EndIf
    \EndFor
    \State $R(\tau) \gets \text{EM}(\text{submitted\_answer}, \text{gold\_answer})$
  \EndFor
  \Statex \hspace{1.5em} \textit{// Phase 3: Policy update}
  \For{each trajectory $\tau \in \mathcal{B}$}
    \If{use\_dense}
      \State $G_t \gets \sum_{k \ge t} s_k$; \quad $b_t \gets \mathrm{Bucket}(a_t,t,\tau)$ \Comment{Reward-to-go and search-rank/terminal bucket}
      \State $A_t \gets \alpha(G_t - \bar{G}_{b_t})$ \Comment{Position-bucketed advantage}
    \Else
      \State $A_t \gets R(\tau) - \bar{R}_{\mathcal{B}}$ \Comment{Outcome advantage}
    \EndIf
  \EndFor
  \State Update $\pi$ by minimizing $\mathcal{L}_{\pi}^{\mathrm{rl}}$ \Comment{Policy-gradient $+\ \lambda_\pi$ KL; Eq.~\ref{eq:pi_rl}}
  \Statex \hspace{1.5em} \textit{// Phase 4: Rubric model co-evolution}
  \State Update $\mu$ by minimizing $\mathcal{L}_{\mu}^{\mathrm{rl}}$ \Comment{Decomposition $+\ \lambda_\mu$ KL; Eq.~\ref{eq:mu_rl}}
\EndFor
\EndFor
\end{algorithmic}
\end{algorithm}

%% file: sections/appendix_runtime.tex
\subsection{Dense-Stage Training Time}
\label{app:runtime}

Table~\ref{tab:dense_runtime} reports the dense-stage wall-clock time up to the best dense-stage checkpoint by dev EM.

\begin{table*}[t]
\centering
\caption{Dense-stage training time to the dense-stage peak-dev-EM checkpoint. EM/F1 are percentages with the $\%$ symbol omitted.}
\label{tab:dense_runtime}
\setlength{\tabcolsep}{6pt}
\renewcommand{\arraystretch}{1.08}
\begin{tabular*}{\textwidth}{@{\extracolsep{\fill}}llcc@{}}
\toprule
Setting & Variant & EM / F1 & Dense time \\
\midrule
\multicolumn{4}{@{}l}{\emph{Main-table ARCO runs}} \\
HotpotQA & Qwen3-4B & 42.80 / 55.07 & 1h 22m \\
HotpotQA & Llama-3.2-3B & 36.40 / 45.58 & 2h 38m \\
2WikiMultiHopQA & Qwen3-4B & 65.20 / 69.38 & 5h 48m \\
2WikiMultiHopQA & Llama-3.2-3B & 59.60 / 63.67 & 1h 52m \\
MuSiQue & Qwen3-4B & 27.40 / 35.55 & 15h 29m \\
MuSiQue & Llama-3.2-3B & 21.60 / 29.70 & 12h 49m \\
\midrule
\multicolumn{4}{@{}l}{\emph{RQ4 rubric-width sweep on HotpotQA / Qwen}} \\
$K{=}1$ & Qwen3-4B & 41.00 / 53.52 & 2h 19m \\
$K{=}3$ & Qwen3-4B & 42.80 / 55.07 & 1h 22m \\
$K{=}5$ & Qwen3-4B & 41.40 / 54.30 & 1h 20m \\
$K{=}7$ & Qwen3-4B & 40.60 / 53.63 & 1h 20m \\
$K{=}9$ & Qwen3-4B & 42.40 / 55.15 & 1h 23m \\
\midrule
\multicolumn{4}{@{}l}{\emph{RQ5 Qwen-family rubric-size sweep on HotpotQA, fixed $\pi$=Qwen3-4B}} \\
$\mu{=}$Qwen3-0.6B & Qwen3-4B & 42.80 / 54.69 & 1h 25m \\
$\mu{=}$Qwen3-1.7B & Qwen3-4B & 42.40 / 54.54 & 1h 27m \\
$\mu{=}$Qwen3-4B & Qwen3-4B & 42.80 / 55.07 & 1h 22m \\
$\mu{=}$Qwen3-8B & Qwen3-4B & 42.00 / 54.32 & 2h 39m \\
\bottomrule
\end{tabular*}
\end{table*}

%% file: sections/appendix_step_signal.tex
\section{Where Does the Step-Level Signal Live?}
\label{app:step_signal}

Section~\ref{sec:rq3_rubric_quality} shows that ARCO's rubrics are bound to the action they score. Here we ask a complementary, quantitative question about the \emph{scores} rather than the rubric text: on which actions does the step score actually separate trajectories that end correct from those that end wrong? For each (dataset, backbone) cell we take the best-dev-EM checkpoint's dev trace---the same rollouts that produce the main-table numbers, so no additional training or inference is needed---and, splitting steps into \texttt{Search} and \texttt{Finish}, we measure two quantities: the \textbf{Gap}, the mean step score on trajectories ending correct (EM${=}1$) minus that on trajectories ending wrong; and the \textbf{AUROC} of the step score as a predictor of final correctness---the probability that a randomly chosen correct trajectory receives a higher step score than a randomly chosen wrong one (chance $0.5$, higher means the score separates correct from wrong better).

\input{sections/search_finish_gap_table.tex}

Table~\ref{tab:search_finish_gap} reveals a consistent asymmetry between the two action types across all six cells: the \texttt{Finish} score is strongly diagnostic of final success (Gap $0.33$--$0.44$, AUROC $0.82$--$0.87$), whereas the \texttt{Search} score is only weakly so (Gap $0.03$--$0.08$, AUROC $0.58$--$0.65$). In other words, the terminal action's score already tells us most of what the outcome will be, while the scores of the intermediate retrieval steps barely do.

We attribute this \texttt{Finish}-versus-\texttt{Search} asymmetry to the single-form process signal noted in our Limitations (\S\ref{sec:limitations}): the only trajectory-level supervision on the scores is the decomposition constraint $\sum_t s_t = R(\tau)$ (Eq.~\ref{eq:mu_rl}), which pins the \emph{sum} of step scores to the outcome but says nothing about how credit should be distributed across steps. We hypothesized that under such a uniform constraint the outcome signal would propagate most cleanly to the step nearest the outcome and dilute along the way to earlier steps: the terminal \texttt{Finish} sits at the shortest path to $R(\tau)$, so its score can align with the outcome almost directly, whereas a \texttt{Search}'s contribution is mediated by every step that follows it. The measured Gap and AUROC confirm this---the score is most reliable at exactly the step whose quality is easiest to read off the answer, and least reliable on the early retrieval decisions where credit assignment would matter most. This points to a concrete direction for future work: replacing the uniform sum constraint with a position-aware or discounted decomposition to push diagnostic signal earlier into the trajectory.

%% file: sections/search_finish_gap_table.tex
% Auto-generated by scripts/rqs/search_finish_gap/compute_gap.py
\begin{table*}[t]
\centering
\caption{Discriminativeness of the ARCO step score on the best-dev-EM checkpoint of each cell. \textbf{Gap} is the mean step score on trajectories that end correct (EM$=1$) minus that on wrong ones; \textbf{AUROC} uses the step score to rank correct vs.\ wrong trajectories (chance $0.5$). Finish actions carry a far stronger signal than Search actions in every cell.}
\label{tab:search_finish_gap}
\setlength{\tabcolsep}{6pt}
\renewcommand{\arraystretch}{1.08}
\begin{tabular*}{\textwidth}{@{\extracolsep{\fill}}ll cc cc@{}}
\toprule
& & \multicolumn{2}{c}{\textbf{Search}} & \multicolumn{2}{c}{\textbf{Finish}} \\
\cmidrule(lr){3-4}\cmidrule(lr){5-6}
Dataset & Backbone & Gap & AUROC & Gap & AUROC \\
\midrule
HotpotQA & Qwen3-4B & 0.071 & 0.604 & 0.329 & 0.816 \\
HotpotQA & Llama-3.2-3B & 0.049 & 0.650 & 0.420 & 0.827 \\
2Wiki & Qwen3-4B & 0.082 & 0.642 & 0.331 & 0.826 \\
2Wiki & Llama-3.2-3B & 0.079 & 0.647 & 0.441 & 0.873 \\
MuSiQue & Qwen3-4B & 0.078 & 0.583 & 0.352 & 0.844 \\
MuSiQue & Llama-3.2-3B & 0.030 & 0.600 & 0.327 & 0.844 \\
\bottomrule
\end{tabular*}
\end{table*}

%% file: sections/appendix_case_figures.tex
\section{Additional Qualitative Examples}
\label{app:qualitative}

\subsection{Rubric Binding (RQ3)}
\label{app:case_figures}

\begin{figure}[ht]
\centering
\includegraphics[
    width=\textwidth,
    trim=25pt 110pt 25pt 110pt,
    clip
]{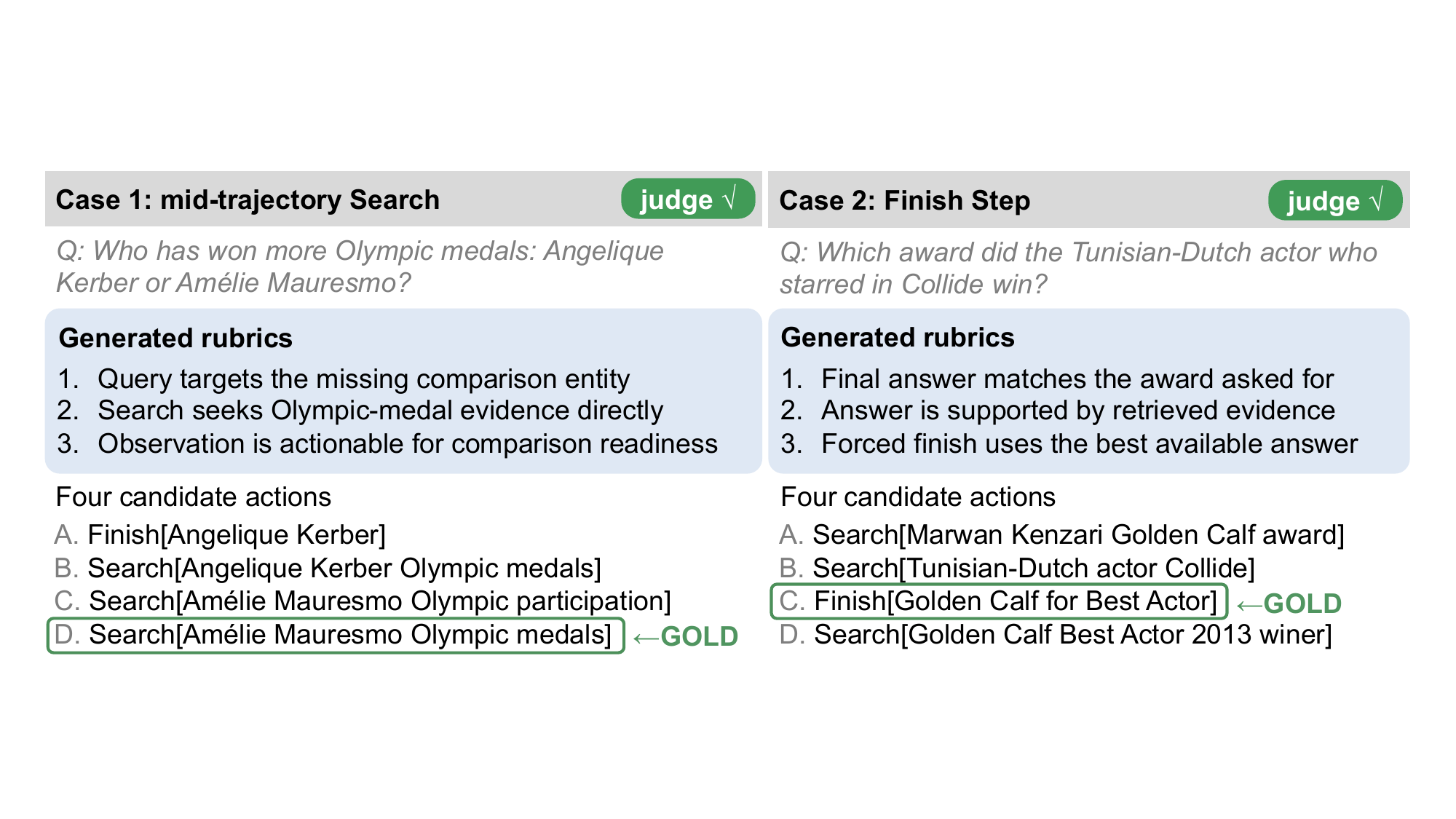}
\vspace{-8pt}
\caption{A HotpotQA dev trajectory under the four-way binding protocol (Qwen3-4B, ARCO epoch 4). The four candidate actions are shared across all three rubric-bearing steps; only the rubric changes, and the judge picks the correct action at each (\checkmark on each card).}
\label{fig:rq3_case_study}
\end{figure}

Figure~\ref{fig:rq3_case_study} complements the RQ3 binding diagnostic with a successful binding case. The candidate actions are reused across rubric-bearing steps, so the judge cannot rely on generic action plausibility or on the question alone---only the rubric changes. The judge selects the correct action at each step, indicating that ARCO's criteria can encode local step intent: different rubrics point to different search or finish decisions even when the distractors are locally plausible.

\begin{figure}[ht]
\centering
\includegraphics[
    width=\textwidth,
    trim=25pt 85pt 3pt 102pt,
    clip
]{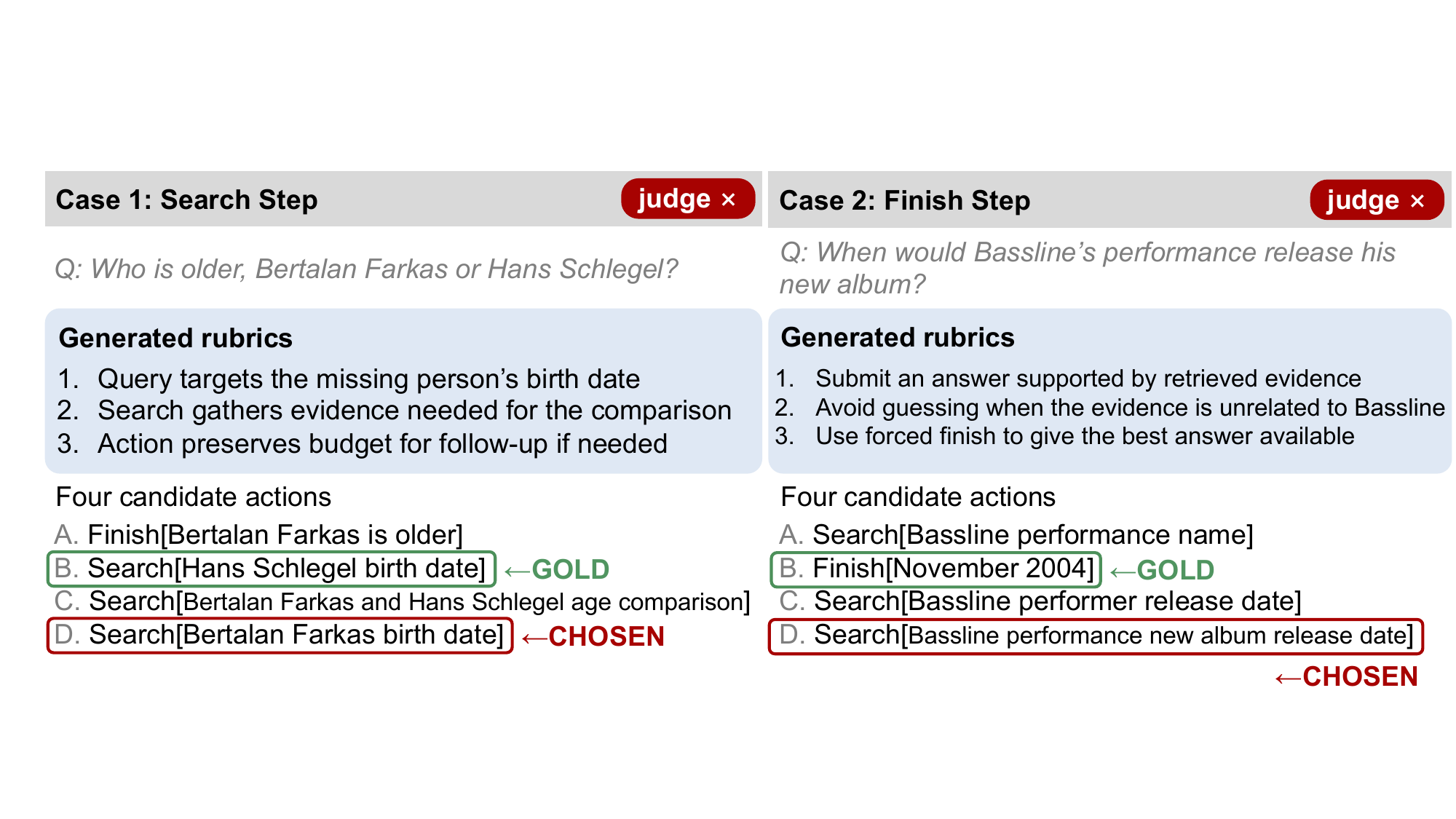}
\vspace{-8pt}
\caption{Two RQ3 rubric-binding failures. In Case 1, the gold action is \texttt{Search[Hans Schlegel birth date]}, but the rubric's phrase ``missing person's birth date'' does not explicitly name Hans Schlegel, so the judge selects the symmetric distractor \texttt{Search[Bertalan Farkas birth date]}. In Case 2, the gold action is a forced \texttt{Finish[November 2004]}, but the rubric emphasizes avoiding unsupported guesses, making an additional release-date search look more aligned.}
\label{fig:rq3_failure_cases}
\end{figure}

Figure~\ref{fig:rq3_failure_cases} shows two representative failure cases. These failures are informative: the rubrics are not nonsensical but describe plausible evaluation dimensions while omitting the contrastive details needed to identify the exact action. In the search-step case, the rubric asks for ``the missing person's birth date''---which correctly describes the role of the next search after one astronaut's birth date has been observed---but does not name Hans Schlegel as the missing entity, so a symmetric distractor that searches for Bertalan Farkas's birth date also looks compatible. In the finish-step case, the rubric asks the judge to submit a supported answer and avoid guessing when evidence is weak; this makes the criteria conservative and less focused on the current action type, so a safer-looking search for the release date is selected instead of the gold \texttt{Finish} action.

These cases suggest a concrete direction for improving rubric generation: step-level rubrics should be more action-discriminative, capturing the characteristic features of the intended action and contrasting it with nearby alternatives. For search steps, criteria should explicitly mention the target entity, relation, or missing evidence item that distinguishes the gold query from previous or alternative searches; for finish steps, they should clearly encode answer readiness and action type, stating that the step is judged as a final submission rather than another evidence-gathering action. Contrastive rubric training with same-trajectory negatives, or prompting the rubric generator to name the distinguishing entity or relation, may therefore improve step binding.

\subsection{Rubric Redundancy (RQ4)}
\label{app:rubric_redundancy}

\begin{figure}[h]
\centering
\includegraphics[
    width=\textwidth,
    trim=0pt 230pt 0pt 0pt,
    clip
]{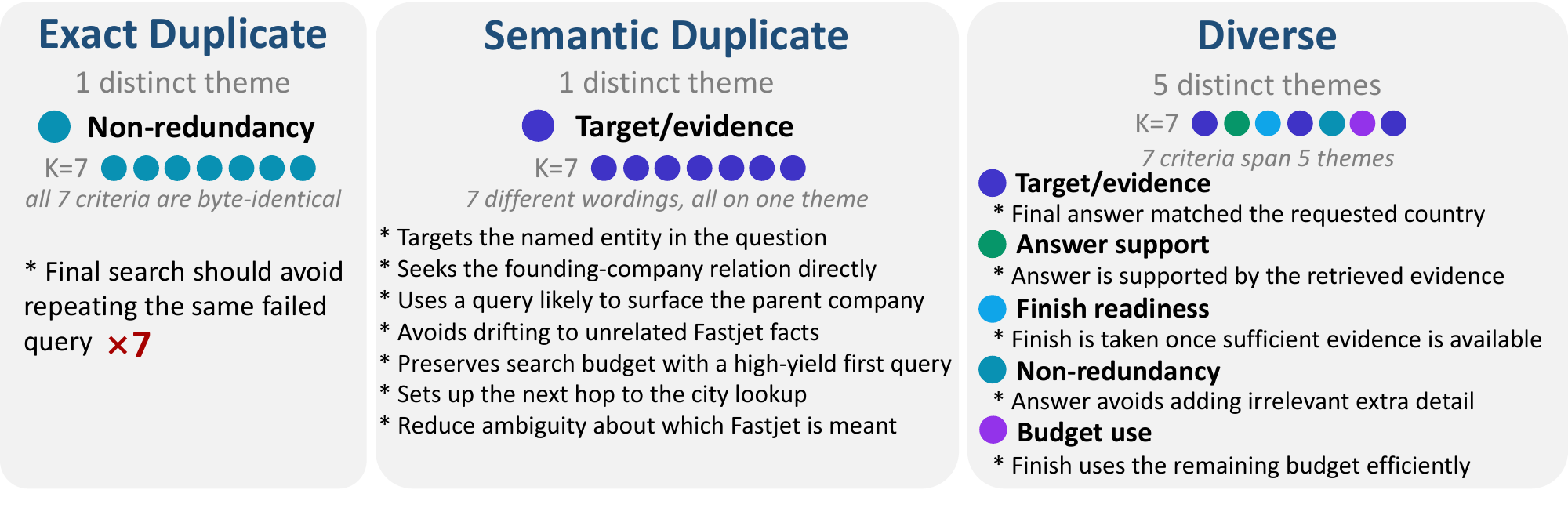}
\caption{Three $K{=}7$ HotpotQA / Qwen dev steps with different redundancy patterns. Colored dots show the theme of each criterion. \emph{Exact duplicate}: 7 byte-identical criteria. \emph{Semantic duplicate}: 7
different wordings all scoring the same theme. \emph{Diverse}: 7 criteria span 5 distinct themes.}
\label{fig:rq4_duplicate_cases}
\end{figure}

Figure~\ref{fig:rq4_duplicate_cases} shows one dev step for each of the three redundancy patterns discussed in \S\ref{sec:rq4_rubric_K}, making concrete how criteria overlap as the rubric width $K$ grows.

%% file: sections/appendix_prompts.tex
\clearpage
\section{Prompts}
\label{app:prompts}

This appendix lists the prompts used by ARCO. The three multi-hop QA datasets share the same prompt structure and differ only in the dataset task description and a few constants such as the search horizon, so we show the HotpotQA version for the policy and rubric prompts. The step-level rubric prompt is used during RL and evaluation, while the trajectory-level rubric prompt is used only during warmup annotation: it asks the teacher to write per-step criteria together with per-step scores that sum to the trajectory's outcome, i.e., satisfy the decomposition constraint $\sum_t s_t = R(\tau)$ (Appendix~\ref{app:warmup}). Finally, we include the two judge prompts for the RQ3 rubric-quality diagnostics (\S\ref{sec:rq3_rubric_quality}): the four-way step-binding judge (\textbf{Bind}) and the step-specificity judge (\textbf{Spec}).

\begin{PromptBox}{Policy Prompt (HotpotQA)}
[SYSTEM]
You are a multi-hop question answering agent. Your goal is to answer complex questions that require gathering information from multiple sources.

Available actions:
- Search[query] : Search for information using a short query.
- Finish[answer] : Submit your final answer.

Guidelines:
- Keep the entire output under 64 tokens.
- You have a limited search budget. Use it to gather the facts you need.
- Do not repeat a search you have already performed.
- Use Finish[answer] as soon as you are confident in the answer.
- For Search[query], use a short query phrase, not a full sentence.
- For yes/no questions, use Finish[yes] or Finish[no].
- For other questions, be concise but complete (e.g. "Owings Mills, Maryland" not just "Maryland").
- Output exactly one action. Do not output multiple actions.
- Output exactly one line in one of these formats, with no angle brackets:
  Search[query]
  Finish[answer]

[USER]
Question: {question}

Previous steps:
{history}

You have {searches_remaining} search(es) remaining. Use them wisely before submitting your answer.

[FORCED FINISH USER]
Question: {question}

Previous steps:
{history}

You have no searches remaining. You must now submit your final answer.
Use Finish[answer] -- do not search again.
\end{PromptBox}
\captionof{figure}{Policy prompt for the HotpotQA agent. The forced-finish variant is used when the search budget is exhausted.}
\label{fig:prompt_policy}
\newpage
\begin{PromptBox}{Rubric System Prompt}
You are an expert evaluator for an agent acting in an interactive task.
Your job is to identify what an action should be judged on in context.

{task_context}

Generate local, context-specific evaluation criteria for the action being evaluated.
These criteria should define what the action ought to be judged on in context.
They may be dimensions, rules, or principles, depending on the task.

The criteria are not a rationale for the score. Do not explain why the model deserves a score. Instead, output the evaluation dimensions, rules, or principles themselves.
Output only the final JSON requested by the user prompt, with no extra text.
\end{PromptBox}
\captionof{figure}{Shared rubric system prompt that defines the evaluator role and output constraints.}
\label{fig:prompt_rubric_system}

\begin{PromptBox}{Step-Level Rubric Prompt}
Question: {question}

Trajectory so far:
{history}

Searches remaining before this action: {searches_remaining}

Action being evaluated: {action}

Generate exactly {K} local criteria for judging this action. The criteria should say what this step ought to be judged on, such as query targeting, planning quality, evidence gathering, answer readiness, or avoiding redundant searches.
If the action is malformed or marked as invalid, judge it as an invalid-action step that wastes budget and should be corrected on the next turn.
If searches remaining is 0, the agent should submit with Finish[answer]; a Search action should be judged as a search-budget violation.
Each criterion should be short and concrete. Keep the full output concise.
Write criterion phrases, not explanations.

Output valid JSON only:
{{"criteria": [{K_schema}]}}
\end{PromptBox}
\captionof{figure}{Step-level rubric prompt that generates local criteria for an individual policy action.}
\label{fig:prompt_step_rubric}
\newpage
\begin{PromptBox}{Warmup Trajectory Rubric Prompt}
Below is a complete trajectory and the final reward it received.
Question: {question}
Trajectory:
{trajectory_text}
Final reward: {reward:.3f}  (0.0 = completely wrong, 1.0 = fully correct)

For each step, write exactly {K} evaluation criteria for that action based on the context available at that point. The criteria should say what this step ought to be judged on, such as query targeting, planning quality, evidence gathering, answer readiness, or avoiding redundant searches. Then assign a score in [-1, +1] for each criterion separately.
If a step is marked as an invalid action or its observation reports an invalid action, judge it as a malformed non-executable step that wastes budget.
If a step is marked as a forced-finish violation, judge it as an invalid final action caused by searching after the search budget was exhausted.
The criteria should be local to the current prefix. Different steps should usually have different criteria.
Do not write a rationale for the score. Write the criteria themselves.
Each criterion should be short and concrete. Keep the full output concise.
Write criterion phrases, not explanations.
Output only the final JSON array.

Scoring guide (range -1 to +1, applies to each individual criterion score):
- Scores are decomposition credits, not standalone quality labels. The final reward is only {reward:.3f}, so do not give large positive scores to many steps unless there are negative scores elsewhere to compensate.
- +0.7 to +1.0 : Very strong positive credit -- use sparingly for a criterion that is essential to the final correct answer.
- +0.2 to +0.6 : Helpful -- this criterion moves the trajectory toward the answer.
- 0.0 : Neutral -- this criterion neither helps nor hurts.
- -0.2 to -0.6 : Harmful or wasteful -- this criterion wastes budget, misses evidence, or adds confusion.
- -0.7 to -1.0 : Strongly harmful -- this criterion causes a wrong final answer or violates the search budget.

CONSTRAINT: Each step must have exactly {K} criteria and exactly {K} scores. The average of the {K} scores is the step's overall score. The sum of all step averages must equal exactly {reward:.3f}. This hard decomposition constraint overrides the local scoring guide: adjust magnitudes, including negative scores when necessary, so the exact sum is {reward:.3f}.
CONSTRAINT: Output exactly one JSON entry for each step in the provided trajectory, in the same order. Do not add extra steps, summaries, or overall entries. Do not omit any step.

Output a JSON array -- one entry per step, in order. Even if there is only one step, wrap it in an array. No markdown fences. No extra text.
[
  {{"step": 0, "criteria": [{K_schema}], "scores": [{K_score_schema}]}},
  {{"step": 1, "criteria": [{K_schema}], "scores": [{K_score_schema}]}},
  ...
]
\end{PromptBox}
\captionof{figure}{Trajectory-level rubric prompt used during warmup annotation to produce criteria and decomposition-compatible pseudo scores.}
\label{fig:prompt_trajectory_rubric}
\newpage
\begin{PromptBox}{RQ3 Step-Binding Judge Prompt}
You are evaluating whether a generated rubric is action-specific.

You are given only the generated rubric and four candidate actions. The question and trajectory context are intentionally hidden.

Your task is to select which candidate action this rubric was most likely generated to evaluate. This condition measures whether the rubric itself points to a concrete action rather than being generic.

Important rules:
- Use only the rubric and candidate actions.
- Do not reward a generally good action unless the rubric specifically matches it.
- Use the content, intent, and stage of each action to determine which one the rubric most directly evaluates.

Generated rubric:
{rubric_bullets}

Candidate actions:
{candidate_actions}

Output JSON only, with no explanation:
{{"choice": "<A|B|C|D>"}}
\end{PromptBox}
\captionof{figure}{RQ3 four-way step-binding judge prompt (\textbf{Bind}). The judge sees only the rubric and four candidate actions; the question and trajectory prefix are hidden, so it can only succeed when the rubric itself points at the gold action.}
\label{fig:prompt_rq3_bind}
\newpage
\begin{PromptBox}{RQ3 Step-Specificity Judge Prompt}
You are judging whether a generated rubric provides local, context-specific criteria for evaluating an agent step.

Question:
{question}

Trajectory before the current step:
{history}

Searches remaining before this action: {searches_remaining_before}
Step index: {step_idx}

Action being evaluated:
{gold_action}

Generated rubric:
{rubric_bullets}

Rate the rubric from 1 to 5:
1 = generic or unrelated to the action
2 = weakly related but mostly generic
3 = partially action-relevant
4 = clearly action-specific and context-aware
5 = highly specific, context-grounded, and directly useful for scoring this step

Output JSON only, with no explanation:
{{"fit": <1-5>}}
\end{PromptBox}
\captionof{figure}{RQ3 step-specificity judge prompt (\textbf{Spec}). The judge sees the question, trajectory prefix, gold action, and the rubric, and rates how specifically the rubric evaluates the gold action on a $1$--$5$ scale.}
\label{fig:prompt_rq3_spec}